\pdfoutput=1

\documentclass[11pt]{article}

\usepackage[final]{acl}

\usepackage{times}
\usepackage{latexsym}

\usepackage{amsmath}
\usepackage{color}
\usepackage{graphicx} 
\usepackage{float} 
\usepackage{subfigure} 
\usepackage{amsmath} 
\usepackage{amssymb} 
\DeclareMathOperator*{\argmax}{argmax} 
\usepackage{multirow}

\usepackage{xcolor}
\definecolor{lightgray}{HTML}{D3D3D3}
\definecolor{darkgray}{HTML}{A9A9A9}

\usepackage{amssymb} 

\usepackage[T1]{fontenc}

\usepackage[utf8]{inputenc}

\usepackage{microtype}
\usepackage{tablefootnote}

\title{Multilingual Knowledge Graph Completion from Pretrained Language Models with Knowledge Constraints}

\author{Ran Song\textsuperscript{1,2}, Shizhu He\textsuperscript{3,4}, Shengxiang Gao\textsuperscript{1,2}, 
\\ \textbf{Li Cai}\textsuperscript{5}, \textbf{Kang Liu\textsuperscript{3,4}}, \textbf{Zhengtao Yu\textsuperscript{1,2} \thanks{ \quad Corresponding author}}, and \textbf{Jun Zhao\textsuperscript{3,4}}
\\ \textsuperscript{1} Faculty of Information Engineering and Automation, \\ Kunming University of  Science and Technology, Kunming, China \\  \textsuperscript{2} Yunnan Key Laboratory of Artificial Intelligence, Kunming, China \\ 
\textsuperscript{3} The Laboratory of Cognition and Decision Intelligence for Complex Systems,\\ Institute of Automation, Chinese Academy of Sciences, Beijing, China \\ \textsuperscript{4} School of Artificial Intelligence, University of Chinese Academy of Science, Beijing, China \\
\textsuperscript{5} Meituan, Beijing, China \\ 
  \texttt{\{song\_ransr\}@163.com},  \texttt{\{shizhu.he,kliu,jzhao\}@nlpr.ia.ac.cn},\\
  \texttt{caili03@meituan.com},  \texttt{\{gaoshengxiang.yn,ztyu\}@hotmail.com}\\  }

\begin{document}
\maketitle
\begin{abstract}
Multilingual Knowledge Graph Completion (mKGC) aim at solving queries like $(h,r,?)$  in different languages by reasoning a tail entity $t$  thus improving multilingual knowledge graphs. Previous studies leverage multilingual pretrained language models (PLMs) and the generative paradigm to achieve mKGC. Although multilingual pretrained language models contain extensive knowledge of different languages, its pretraining tasks cannot be directly aligned with the mKGC tasks. Moreover, the majority of KGs and PLMs currently available exhibit a pronounced English-centric bias. This makes it difficult for mKGC to achieve good results, particularly in the context of low-resource languages. To overcome previous problems, this paper introduces global and local knowledge constraints for mKGC. The former is used to constrain the reasoning of answer entities, while the latter is used to enhance the representation of query contexts. The proposed method makes the pretrained model better adapt to the mKGC task. Experimental results on public datasets  demonstrate that our method outperforms the previous SOTA on Hits@1 and Hits@10 by an average of 12.32\% and 16.03\%, which indicates that our proposed method has significant enhancement on mKGC.
\end{abstract}

\section{Introduction}
\label{intro}

Knowledge graphs are collections of entities and facts, and utilized as a valuable resource in a variety of natural language processing (NLP) tasks, such as Question Answering and Recommender Systems ~\cite{shah2019kvqa, du2021knowledge, wang2019knowledge}. The language-specific nature of many NLP tasks necessitates to consider the  knowledge expressed in a particular language. For example, multilingual question answering needs multilingual knowledge graphs ~\cite{zhou2021improving}. The utilization of multilingual knowledge graphs (mKGs) with a vast amount of knowledge in multiple languages, such as DBpedia~\cite{lehmann2015dbpedia}, Wikidata~\cite{vrandevcic2014wikidata}, can be advantageous in plenty of NLP tasks~\cite{zhou2021improving, fang2022leveraging}.

\begin{figure}
\centering 
\includegraphics[width=7.5CM]{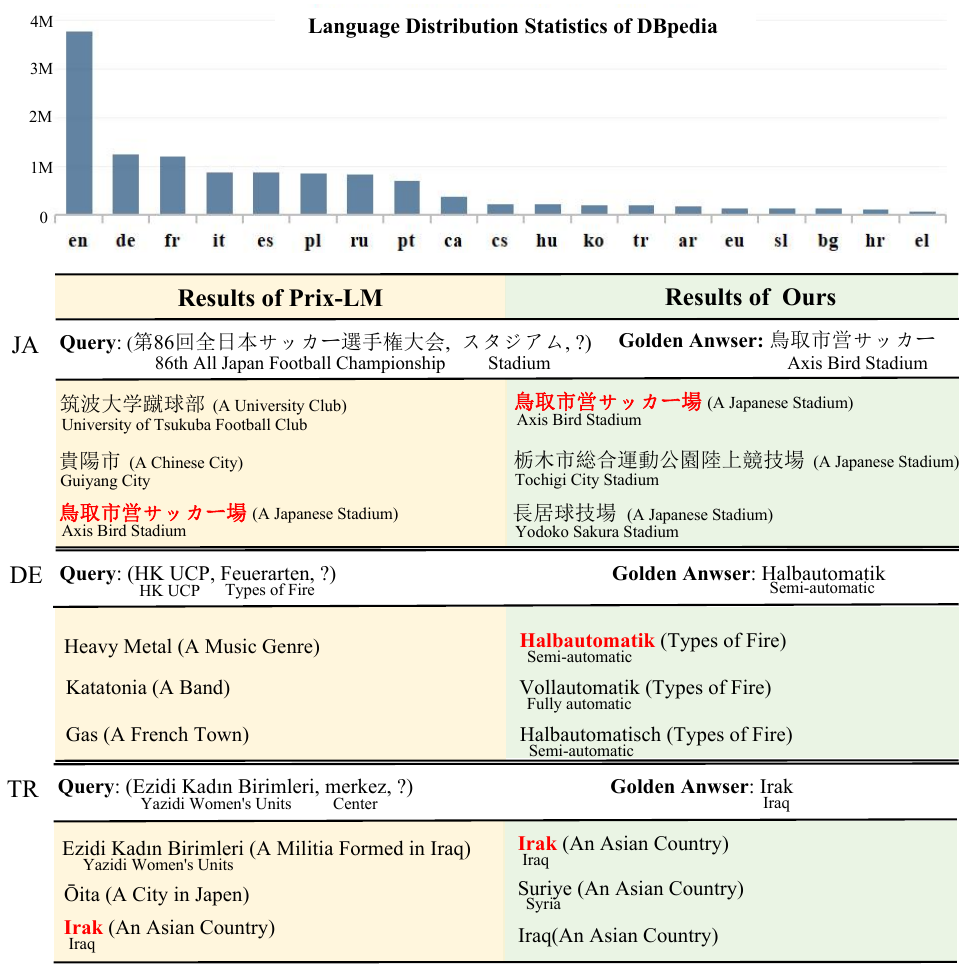}
\caption{The top part introduces unbalance language distribution for DBpedia. The low part shows the sampling comparison results of Prix-LM model and our method. The type of prediction entity and the correct answer are shown in brackets and \textbf{\textcolor{red}{red}} font, respectively. Our approach exhibits superior consistency and accuracy in generating answers.}
\label{Fig.main1}
\end{figure}

There is a significant amount of potential facts that have not been captured in current knowledge graphs, resulting in their incompleteness ~\cite{chen2020knowledge}.
To address this issue, various studies have proposed for Knowledge Graph Completion (KGC) to automatically discovery potential facts through observed facts~\cite{bordes2013translating}, rules~\cite{meilicke2019anytime} and language models~\cite{lv2022pre}. 

In fact, as shown in Figure~\ref{Fig.main1}, there is more English-centric knowledge than other languages, so that it is difficult to leverage knowledge graphs on non-English tasks. For example, English-centric commonsense reasoning tasks obtain better development and performance than other languages ~\cite{lin2021common}. And the knowledge coverage of non-English knowledge graphs is even worse, it will poses challenges for traditional KGC methods to achieve superior performance.

Nowadays, pretrained language models (PLMs) learn various knowledge modeling  capabilities~\cite{petroni2019language, jiang2020can} from massive unlabeled data. 
And most studies have demonstrated that the knowledge contained within PLMs can significantly improve the performance of downstream tasks~\cite{li2021pretrained,lin2021pretrained}. 
Most recently, Prix-LM \cite{zhou-etal-2022-prix} approached mKGC as an end-to-end generative task using multilingual PLMs. For example, for predicting the missing entity of the query \textit{(86th All Japan Football Championship, Stadium, ?)}  (see Figure~\ref{Fig.main1}), Prix-LM converts the query into a sequence 
with pre-defined template, which is then processed by an encoder to generate a query representation. The decoder then uses this representation to generate the final answer \textit{Axis Bird Stadium}.  

Despite the successes achieved through the combination of PLMs and the generative paradigm, there remain limitations for mKGC. 
On the one hand, the gap between the pretraining task and the KGC task may contribute to the limitations. It arise that the answers generated by Prix-LM are ambiguous in type.  On the other hand, languages and tokens that occur more frequently in the pretraining data have richer representations. Linguistic bias for KGs and PLMs would arise that entities in low-resource languages are difficult to be represented, resulting answer incorrect. As illustrated in Figure~\ref{Fig.main1}, the query \textit{(86th All Japan Football Championship, stadium, ?)} expects a response of the type \textit{stadium}, but the top-ranked answers from Prix-LM are diverse, and the top answer is incorrect. 

We argue that the incorporation of knowledge constraints into the generation process  can increase PLMs suitability for mKGC tasks.  We categorize knowledge effective for mKGC into global and local knowledge.
Global Knowledge limit the types of answers based on building the relationship of entity and relation representations. This helps to ensure that the generated answers are semantically and logically consistent with the intent of query.
On the other hand, local knowledge in PLMs can enhance the ability to comprehend the interconnections between the sub-tokens within the query.  This helps the model to better understand the context of query and generate more accurate answers.
Incorporating knowledge constraints into the generative process brings two advantages for mKGC: 1) It makes PLMs to better adapt to mKGC task. 2) It enables PLMs to learn more effective representations from low-resource data. 

In this paper, we propose  to incorporate the global and local knowledge into the answer generation process through two knowledgeable tasks. To learn global knowledge, special tokens ([H],[R],[T]) are introduced as semantic representations of head entity, relation, and tail entity in a triple. A scoring function measures the plausibility of the resulting facts, such  as $||h_{[H]}+h_{[R]} - h_{[T]}||_{L_{1/2}}$. Since the same special token is used in each triple in different languages, trained models are able to learn knowledge reasoning ability beyond language boundaries. To capture local knowledge, we consider the representation of answer and each word of query as two separate distributions $P(H_q)$ and $P(H_{[T]})$, and then use an estimator to estimate and maximize the mutual information between them $I(H_q;H_{[T]})$. The local knowledge serves to augment the query representations for trained model through the utilization of minimal amounts of data. The experimental results on seven  language knowledge graph from DBpedia show that our proposed method achieves significant improvement as compared to Prix-LM and translated-based methods. We publicize the dataset and code of our work at \url{https://github.com/Maxpa1n/gcplm-kgc}.

\begin{figure*}[htp]
\centering
\includegraphics[height=7cm]{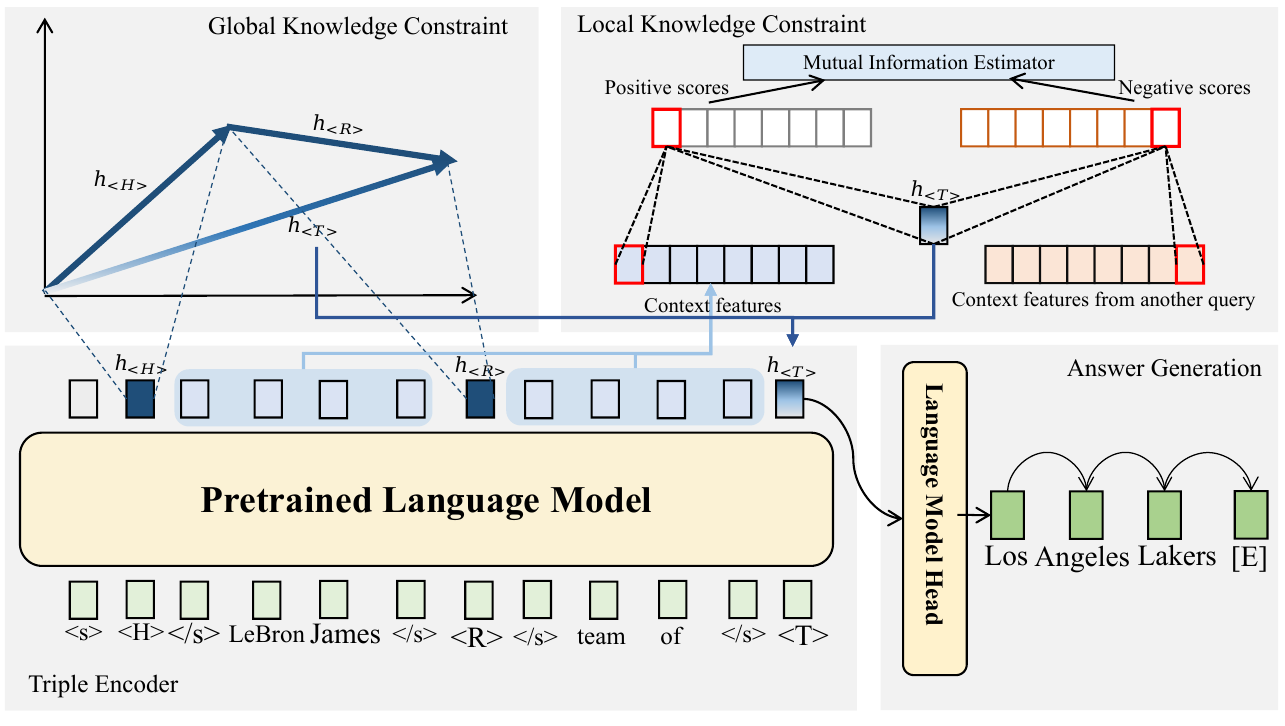}
\caption{This figure illustrates the architecture of the complete model, which is composed of four main components: a query encoder, a global knowledge constraint, a local knowledge constraint, and an answer generation module. The global knowledge learn from representations of head and relation (\textcolor[RGB]{31,78,121}{navy blue}). The local knowledge learn from representations of query words (\textcolor[RGB]{218,227,243}{light blue}). We use different colors to represent entities and relation in each module for a triple.}
\label{fig:picture001}
\end{figure*}

In short, our main contributions are as follows:
\begin{itemize}
  \item 
    We attempt to utilize diverse knowledge constraints to enhance the performance of PLM-based mKGC. It effectively addresses the inconsistency of PLM and mKGC task, and alleviates language and data bias from PLMs and KGs.
  \item 
    Our proposed method can enrich query representation and enhance answer generation by introducing global knowledge constraints for entity placeholders and mutual information constraints for other contextual symbols.

  \item 
     Our proposed method outperforms the Prix-LM~\cite{zhou-etal-2022-prix} in both mKGC and cross-lingual entity alignment, as shown by experiments on a public dataset. The performance of our method on Hits@1, Hits@3, and Hits@10 shows an average improvement of 12.32\%, 11.39\%, and 16.03\%, respectively.
\end{itemize}

\section{Basic Model}
A knowledge graph $\mathcal{G}=(\mathcal{R},\mathcal{E})$ is a collection of connected information about entities, often represented using triples $(h,r,t)$ where $r \in \mathcal{R}$ is relation and $h,t \in \mathcal{E}$ are entities. Prix-LM is an important work of mKGC and is also used as the basic model in this paper. Prix-LM transfer link prediction from discriminative task to generative task for mKGC. The goal of mKGC is to generate the missing tail entity, which may contain multiple tokens, for the query $(h,r,?)$ of different languages.  The use of template is employed as a means of transforming queries into textual sequences that can be encoded by PLMs. The template includes special tokens, which serve to identify the specific role of each element within the query triple:

\noindent \setlength{\fboxsep}{10pt}
\colorbox{lightgray}{
\begin{minipage}{6.5cm}
<s>[H]$X_h$</s></s>[R]$X_r$</s></s>[T]$X_t$[E]
\end{minipage}
}

\noindent where <s> is beginning token of sentence and </s> is the separator, both are applied in PLMs, as known as [CLS] and [SPE]. [H], [R] and [E] are additional special tokens for the representation of head, relation and tail. [E] is the end-of-sequence token. $X_h \in \{\mathbf{x}_1^h, \mathbf{x}_2^h, \mathbf{x}_3^h,...,\mathbf{x}_n^h\}$ are text words of head entity, $X_r$ and $X_t$ in the same way.

The training goal is to generate the tail entity $X_t$ by giving the sequence containing the head entity $X_h$ and relation $X_r$. For example, for the query  \textit{(LeBron James, team member of, ?)}, the constructed sequence is <s>[H] \textit{LeBron James}</s></s>[R] \textit{team member of}</s></s>[T], and the target of mKGC is generate \textit{Los Angeles Lakers} [E]. The process is as follows:
\begin{equation}
\begin{split}
P_{\theta}(X_{t}|X_{h},X_{r}) = \prod_{x_i \in X_t}^{x_i}p(x_i|x_{<i},\theta)
\end{split}
\end{equation}
where $\theta$ is the pretrained model parameter. According to the mechanism of causal language model, the probability of $i$-th token depend on previous token representation $\mathbf{h}_{i-1}$:  
\begin{equation} \label{softmax}
\begin{split}
p(x_i|X_{<i}) = \mathrm{softmax}(\mathbf{W}\mathbf{h}_{i-1}) 
\end{split}
\end{equation}
where $W$ is causal language model decoder from PLMs.

The utilization of PLMs for generating answers directly can be subject to language bias, resulting in ambiguous and incorrect answers. The representation of the special token $[T]$ is a crucial factor in determining the quality of the generated answers. To improve the representation of the $[T]$ token, we have implemented two supplementary strategies aimed at incorporating additional knowledge into its representation.

\section{The Proposed Model}
In this section, we describe the components of our proposed approach.  The architecture of the model is depicted in Figure ~\ref{fig:picture001}. Our approach comprises four key components: a query encoder, a global knowledge constraint, a local knowledge constraint, and an answer generation module. These components operate in tandem to generate accurate and coherent answers for given queries.

\subsection{Triple Encoder}
\begin{figure}[htp]
\centering
\includegraphics[height=5cm]{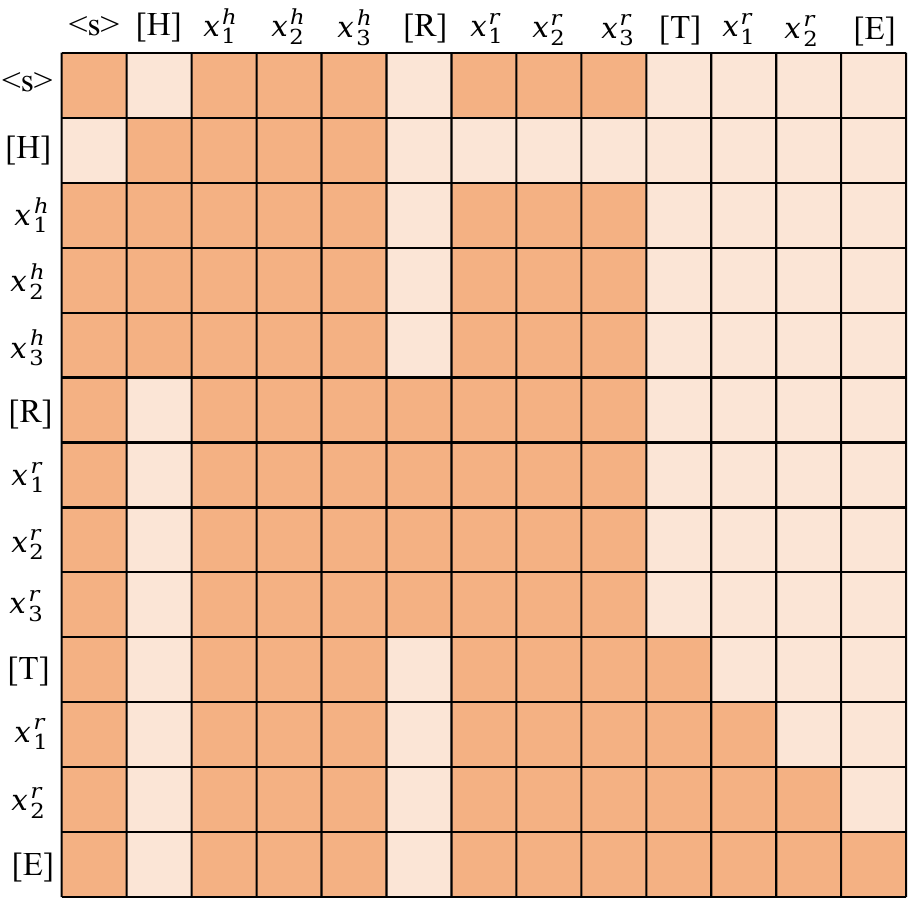}
\caption{The operation mechanism of mask matrix during training process. The darker squares indicate that attention is allowed, while the lighter squares indicate that attention is suppressed. }
\label{fig:mask}
\end{figure}

We leverage the PLM to encode the triple and an attention mask to control the access to each subtoken in the sequence during training process. We use previous template to convert a triple $(h, r, t)$ to a sequence $S_{(h,r,t)} \in \{X_h, X_r, X_t, X_a\}$, and $X_a$ is special token. The attention mask mechanism allows the query sequence to be seen as the source text and the answer entity as the target text. The process as following:
\begin{equation}
\begin{split}
\mathrm{PLM}(S_{(h,r,t)}) = H
\end{split}
\end{equation}
where hidden representation of triple is $ H \in \{\mathbf{h}^{[H]},\mathbf{h}_1^h,..,\mathbf{h}^{[R]},\mathbf{h}_1^r,...,\mathbf{h}^{[T]},\mathbf{h}_1^t,...,\mathbf{h}^{[E]}\}$. The attention mask is a matrix that specifies whether each subtoken should be attended or ignored, as illustrated in Figure ~\ref{fig:mask}. By making special tokens only visible to their own subtokens, model can effectively separate each role in a triple. And the mask matrix $\mathbf{M}$ add in attention score calculated by query $\mathbf{Q}$, key $\mathbf{K}$, value $\mathbf{V}$:
\begin{equation}
\begin{split}
\mathbf{M} = \begin{cases}
0, & \mathrm{allow \ to \ attend}\\
-\infty, & \mathrm{prevent\ from \ attending}\\
\end{cases}
\end{split}
\end{equation}
\begin{equation}
\begin{split}
\mathbf{A} = \mathrm{softmax}(\frac{\mathbf{Q}\mathbf{K}^T}{\sqrt{d}}+\mathbf{M})\mathbf{V}
\end{split}
\end{equation}
where $\mathbf{Q}$, $\mathbf{K}$, $\mathbf{V}$ $\in \mathbb{R}^{l \times d}$, $l$ is length of the input sequence, and $d$ is the hidden size. 

\subsection{Global Knowledge Constraint}
To bridge the gap between the pretraining task and the KGC task, we introduce  the global knowledge build  logical relationship between entities.  Unlike previous approaches such as Prix-LM, our method does not rely on cross-lingual links for equivalent entities to learn shared knowledge in different languages.  Instead, shared knowledge between languages is learned through the global knowledge constraint, which is inspired by embedding-based methods. We leverage the TransE framework in our model, and methods such as CompleX, RotatE are also applicable. The goal of the global knowledge constraint is to represent entities and relation in a semantic space and enforce the translational principle: $h+r \approx t$: 
\begin{equation}
\begin{split}
\Vert \mathbf{h}_{[H]}+\mathbf{h}_{[R]} \Vert = \Vert \mathbf{h}_{[T]} \Vert
\end{split}
\end{equation}
where $\mathbf{h}_{[H]}$, $\mathbf{h}_{[R]}$, $\mathbf{h}_{[T]}$ are special tokens representation, and $\Vert . \Vert$ is L1 norm. And a triple global knowledge score is described by:
\begin{equation}
\begin{split}
score(h,r,t) =  \Vert \mathbf{h}_{[H]}+\mathbf{h}_{[R]}  -  \mathbf{h}_{[T]} \Vert
\end{split}
\end{equation}
\noindent We use the same special tokens for different languages. The following loss function is used to optimize the model.

\begin{equation}
\begin{split}
\mathcal{L}_p = \sum_{g_j \in G}^{g_j} \sum_{(h,r,t)_i \in g_j}^{(h,r,t)_i}(score(h_i,r_i,t_i)+\gamma)
\end{split}
\end{equation}
where $G$ is all language knowledge graphs set, and $\gamma$ is correction factor. 

\subsection{Local Knowledge Constraint}
The local knowledge enables the model to learn more accurately for generated answers with low-resource data. Therefore, we consider establishing the connection between query and answer in a triple. Specifically, we view the the representation of query words $H_q $ and tail entity $H_{[T]}$ as two distributions and maximizing the mutual information between them $I(H_q, H_{[T]})$. The theoretical foundation for this idea is provided by MIEN \cite{belghazi2018mutual}, which demonstrates that mutual information follows a parametric lower-bound:
\begin{equation}
\begin{split}
I(H_q;H_{[T]}) \geq \hat{I}_{\theta}(H_q;H_{[T]})
\end{split}
\end{equation}

Inspired from previous Mutual Information Maximization ~\cite{tschannen2019mutual,zhang2020unsupervised} (MIM) method in unsupervised learning, we take the local features, represented by $H_q$, and the global features, represented by $H_{[T]}$, as the inputs for MIM.  Benefit from the mask mechanism and PLM's powerful learning capability, we do not strictly distinguish the parameter of encoder and decoder different from previous works. In this work, we select a Jensen-Shannon MI estimator to parameterize Mutual Information:
\begin{equation}
\begin{split}
\hat{I}^{(JSD)}_{\theta}(H_q, H_{[T]}) := \\
E_\mathbb{P}[-sp(T_{\theta}(H_q,H_{[T]}))] \\
-E_{\mathbb{P} X \widetilde{\mathbb{P}}}[sp(T_{\theta}(H_q^{'},H_{[T]}))]
\end{split}
\end{equation}
where $H_q \in \{\mathbf{h}_1^h,\dots,\mathbf{h}_m^h ,\mathbf{h}_1^r,\dots,\mathbf{h}_n^r\} $ is query words representation, $m$ is head entity length, $n$ is relation length. $H_{[T]} \in \{\mathbf{h}_{[T]}\}$ is tail entity representation. $T_{\theta}$ is a discriminator function support by the PLM parameters. $H_q^{'}$ is representation sampled from other query in the same min batch. And $\mathbb{P} = \widetilde{\mathbb{P}}$ make guarantee the expectation easy to calculated. $sp(x)=log(1+e^x)$ is the softplus activation function. The learning object is to make PLM estimate and maximize the Mutual Information:
\begin{equation}
\begin{split}
\theta = \argmax_{\theta} \frac{1}{|G|}\sum_{b_j \in G}^{b_j} \hat {I}^{(JSD)}_{\theta}(H^j_q, H^j_{[T]})
\end{split}
\end{equation}
where $b_j$ is mini batch from training dataset. To optimize model by gradient descent, we set loss function as following:
\begin{equation}
\begin{split}
\mathcal{L}_E= \sum_{b_j \in G}^{b_j} (E^j_{ \widetilde{\mathbb{P}}}-E^j_\mathbb{P})
\end{split}
\end{equation}
where the $E^j_\mathbb{P}$ is expectation for query and tail entity. The local knowledge constraint within PLM enhance its capacity to obtain rich representations of queries and tail entities, particularly in situations where training data is limited.

\subsection{Answer Generation Module}
Follow the paradigm that given a serialized query and generate answer token, we use the casual language model with PLM. The generation loss function as Cross Entropy Loss function:
\begin{equation}
\begin{split}
\mathcal{L}_G= \sum_{(h,r,t)_i \in G}^{i} \sum_{ x_j \in X_i^t}^{x_j} x_jlog(f(x_{<j}))
\end{split}
\end{equation}
where the $f( \cdot )$ is like Formula~\ref{softmax}, $x_j$ is subtoken of tail entity. 

In training process, the model would generate answer with global and local knowledge, we define the loss for model as:
\begin{equation}
\begin{split}
\mathcal{L}= \mathcal{L}_G +\alpha\mathcal{L}_P+ \beta\mathcal{L}_E
\end{split}
\end{equation}
where $\alpha$ and $\beta$ are hyperparameter. The mask mechanism achieved that all subtokens of tail entity be trained in one round.  

\subsection{Inference}
During the inference phase of our model, we utilize an autoregressive approach to generate the tokens of tail entity for given query. This autoregressive approach involves predicting the next token based on the previous tokens. The query $(h,r,?)$ be transferred  to a sequence $X_q$ and generating the answer entity by trained model. The process as following:
\begin{equation}
\begin{split}
x_{i} =  \argmax_{x_{i}}P(x_i|X_q \cap x_1\dots \cap x_{i-1})
\end{split}
\end{equation}
where $x_i \in X_t$. Additionally, we assume a closed-world setting and utilize constrained beam search to restrict the final output to a predefined set of possibilities, in order to ensure the validity of the generated answer.

\begin{table*}
\small
\setlength{\tabcolsep}{3.9mm}{
\begin{tabular}{llrrrrrrrr}
\hline
    & MODEL   & \multicolumn{1}{r}{DE} & \multicolumn{1}{r}{FI} & \multicolumn{1}{r}{FR} & \multicolumn{1}{r}{HU} & \multicolumn{1}{r}{IT} & \multicolumn{1}{r}{JA} & \multicolumn{1}{r}{TR} & \multicolumn{1}{l}{AVG} \\ \hline
\multirow{7}{*}{\rotatebox{90}{Hits@1}}      & TransE     & 0.00   & 0.01   & 0.02   & 0.03   & 0.04   & 0.02  & 0.06   & 0.02   \\
    & ComplEx    & 4.09   & 2.45   & 2.50   & 3.28   & 2.87   & 2.41   & 1.00   & 2.65   \\
    & RotatE     & 6.72   & 5.87   & 8.40   & 16.27  & 6.91   & 6.21   & 6.85   & 8.17   \\
    \cline{2-10}
    & Prix-LM (Single)     & 12.86  & 19.81  & 18.01  & 28.72  & 16.21  & 19.81  & 23.79  & 19.88  \\
    & Prix-LM & 14.32  & 18.78  & 16.47  & 29.68  & 14.32  & 18.19  & 21.57  & 19.04  \\
    & \textbf{Ours} & \textbf{17.54} & \textbf{20.74} & \textbf{18.34} & \textbf{30.91} & \textbf{14.98} & \textbf{22.05} & \textbf{25.20} & \textbf{21.39} \\ \hline
\multirow{7}{*}{\rotatebox{90}{Hits@3}}      & TransE     & 6.14   & 6.54   & 6.60   & 14.91  & 5.95   & 7.22   & 8.20   & 7.93   \\
    & ComplEx    & 8.47   & 5.28   & 5.19   & 6.70   & 4.31   & 4.68   & 2.11   & 5.24   \\
    & RotatE     & 10.52  & 7.42   & 14.62  & 21.75  & 12.11  & 9.75   & 11.29  & 12.49  \\
    \cline{2-10}
    & Prix-LM (Single)    & 23.09  & 28.75  & 24.75  & 38.44  & 25.32  & 29.02  & 33.05  & 28.91  \\
    & Prix-LM & 23.68  & 29.54  & 23.15  & 39.80  & 25.46  & 27.01  & 31.45  & 28.58  \\
    & \textbf{Ours} & \textbf{30.40} & \textbf{29.74} & \textbf{26.36} & \textbf{44.18} & \textbf{27.03} & \textbf{30.79} & \textbf{35.48} & \textbf{31.99} \\ \hline
\multirow{7}{*}{\rotatebox{90}{Hits@10}}     & TransE     & 17.54  & 17.80  & 15.26  & 29.00  & 14.16  & 20.65  & 19.35  & 19.10  \\
    & ComplEx    & 9.35   & 8.21   & 8.91   & 16.96  & 8.76   & 8.23   & 5.24   & 9.38  \\
    & RotatE     & 14.61  & 8.61   & 19.49  & 28.31  & 18.48  & 14.44  & 17.13  & 17.29  \\
    \cline{2-10}
    & Prix-LM (Single)     & 33.82  & 38.91  & 34.04  & 47.31  & 36.61  & 38.81  & 38.50  & 38.28  \\
    & Prix-LM & 33.91  & 41.29  & 32.25  & 46.23  & 35.18  & 36.12  & 37.50  & 37.49  \\
    & \textbf{Ours} & \textbf{41.81} & \textbf{43.44} & \textbf{35.15} & \textbf{58.00} & \textbf{39.15} & \textbf{42.45} & \textbf{44.55} & \textbf{43.50} \\ \hline
\end{tabular}}
    \caption{In this table, the results of seven language-specific  knowledge graph completion (KGC) tasks are presented. The embedding-based methods, including TransE, complEx, and RotatE, were implemented using the OpenKE framework~\cite{han2018openke}. The results for these methods were obtained by training separate knowledge graphs for each language. The Single make the  monolingual version, which is trained independently for each language. The numbers in \textbf{bold} represent the best results among the methods and languages considered.}
    \label{tab:main}
\end{table*}

\section{Experiments}
In this section, we evaluate the effectiveness of our approach on tasks related to mKGC and Entity Alignment for mKGs. To further understand the role of the various knowledge-gathering strategies in our method, we also conduct ablation experiments. Additionally, we provide case studies to demonstrate the superior performance of our method on specific examples. These experiments and analyses provide insight into the strengths and limitations of our approach for addressing challenges in mKGC for sparse knowledge graphs.

\begin{table}
\tiny
\setlength{\tabcolsep}{0.7mm}{
\begin{tabular}{llrrrrrrrr}
\hline
    &    & \multicolumn{1}{r}{DE} & \multicolumn{1}{r}{FI} & \multicolumn{1}{r}{FR} & \multicolumn{1}{r}{HU} & \multicolumn{1}{r}{IT} & \multicolumn{1}{r}{JA} & \multicolumn{1}{r}{TR} & \multicolumn{1}{r}{AVG} \\ \hline
\multirow{3}{*}{Training}   & Entity     & 39,842 & 36,892 & 106,955& 27,765 & 86,988 & 68,279 & 29,120 & 56,549  \\
    & Relation   & 1,544  & 945    & 2,358  & 999    & 1,539  & 2,542  & 1,008  & 1,562   \\
    & Triple     & 27,014 & 28,040 & 83,940 & 24,193 & 66,904 & 50,164 & 24,013 & 43,467  \\ \hline
\multirow{3}{*}{Validation} & Entity     & 501    & 766    & 2,452   & 988    & 2,240   & 1,206   & 749    & 1,272   \\
    & Relation   & 122    & 142    & 362    & 142    & 257    & 303    & 101    & 204     \\
    & Triple     & 264    & 435    & 1,407  & 614    & 1,286  & 671    & 432    & 730     \\ \hline
\multirow{3}{*}{Testing}    & Entity     & 649    & 916    & 2,687  & 1,154  & 2,499  & 1,414  & 822    & 1,449   \\
    & Relation   & 135    & 147    & 377    & 154    & 271    & 322    & 95     & 214     \\
    & Triple     & 342    & 511    & 1,559  & 731    & 1,461  & 789    & 496    & 841     \\ \hline
\multicolumn{2}{l}{T/E Ratio}  & 0.69   & 0.79   & 0.81   & 0.92   & 0.80   & 0.76   & 0.76   & 0.79    \\ \hline
\end{tabular}}
    \caption{The table show the statistics of multi language knowledge graph dataset. The T/E Ratio is equal to the number of triples divided by the number of entities.}
    \label{tab:number}
\end{table}

\subsection{Datasets and Evaluation Metrics}
To evaluate our method, we utilize the Link Prediction dataset provided by Prix-LM~\cite{zhou-etal-2022-prix} and split it by the closed-world setting. The dataset consists of data from DBpedia, a large multilingual knowledge graph, and the amount of data is shown in Table~\ref{tab:number}. We ensure that entities and relations appearing in the validation and test sets are included in the training set. We introduce a ratio between entities and triples as a measure of the knowledge density of the dataset. This ratio has a lower bound of 0.5, which indicates that there are no cross-links between triples. The ratio of our dataset is much lower than that of publicly available datasets. The evaluation metrics we use are standard Hits@1, Hits@3, and Hits@10, which are commonly used in the evaluation of KGC methods.

\subsection{Implementation Details}
In our experiments, we used XLM-R (Base) as the base pre-trained language model and did not introduce any additional parameters beyond those provided by XLM-R. The model was implemented using the Huggingface Transformers library ~\cite{wolf-etal-2020-transformers} and the hyperparameters $\alpha$ and $\beta$ were set to 0.001 and 0.005. The learning rate and batch size were selected from the sets \{4e-5, 5e-5\} and \{128, 256\}. And the maximum length of a triple sequence was 35. The model was trained using a single Nvidia RTX 3039 GPU.

\subsection{Multilingual Knowledge Graph Completion}
Our method for mKGC was compared to various embedding-based methods and Prix-LM on seven languages KG, as shown in Table~\ref{tab:main}. The results show that our method outperformed Prix-LM on the metrics of Hits@1, Hits@3, and Hits@10, with average improvements of 12.32\%, 11.39\%, and 16.03\%, respectively. These improvements suggest that the integration of both global and local knowledge significantly enhances the effectiveness of the mKGC task, leading to a higher ability to accurately predict missing triple in KG. 

It is worth noting that the low knowledge density in the training set can hinder the performance of traditional embedding-based methods, which rely on the presence of sufficient training data to learn meaningful relationships between entities. In contrast, the use of PLMs, as employed in our method, can effectively address the issue of data sparsity and still achieve notable impact on performance. Overall, these results demonstrate the effectiveness of our approach in comparison to the use of PLMs alone for mKGC.

\begin{table}
\small
\centering
\setlength{\tabcolsep}{2.8mm}{
\begin{tabular}{clllll}
\hline
\multicolumn{1}{l}{}     & model   & zh-km & zh-th & zh-lo & zh-my \\ \hline
\multirow{2}{*}{\rotatebox{90}{H1}}  &  Prix-LM    & 65.02 & 22.42 & 62.15 & 41.56 \\
 & Ours  & \textbf{67.25} & \textbf{24.07} & 62.15 & \textbf{43.49} \\
\hline
\multirow{2}{*}{\rotatebox{90}{H3}} & Prix-LM    & 68.37 & 26.87 & 64.27 & 46.44 \\
&  Ours   & \textbf{69.23} & \textbf{27.65} & \textbf{64.80} & \textbf{47.65} \\
\hline          
\multirow{2}{*}{\rotatebox{90}{H10}}   & Prix-LM    & 70.15 & \textbf{30.50} & 66.12 & 49.08 \\ 
& Ours & 70.15 & 30.35 & \textbf{66.27} & \textbf{50.73} \\ \hline
\end{tabular}}
\caption{This table presents the results of entity alignment tasks for low-resource languages. The parallel entity pairs were obtained from Wikidata~\cite{vrandevcic2014wikidata}. We transformed the entity alignment into a KGC task by augmenting the knowledge graph with additional edges representing the linguistic relations between the entity pairs. }
    \label{tab:ea}
\end{table}

\begin{figure*}[htp]
\centering
\includegraphics[height=6cm, width=15.7cm]{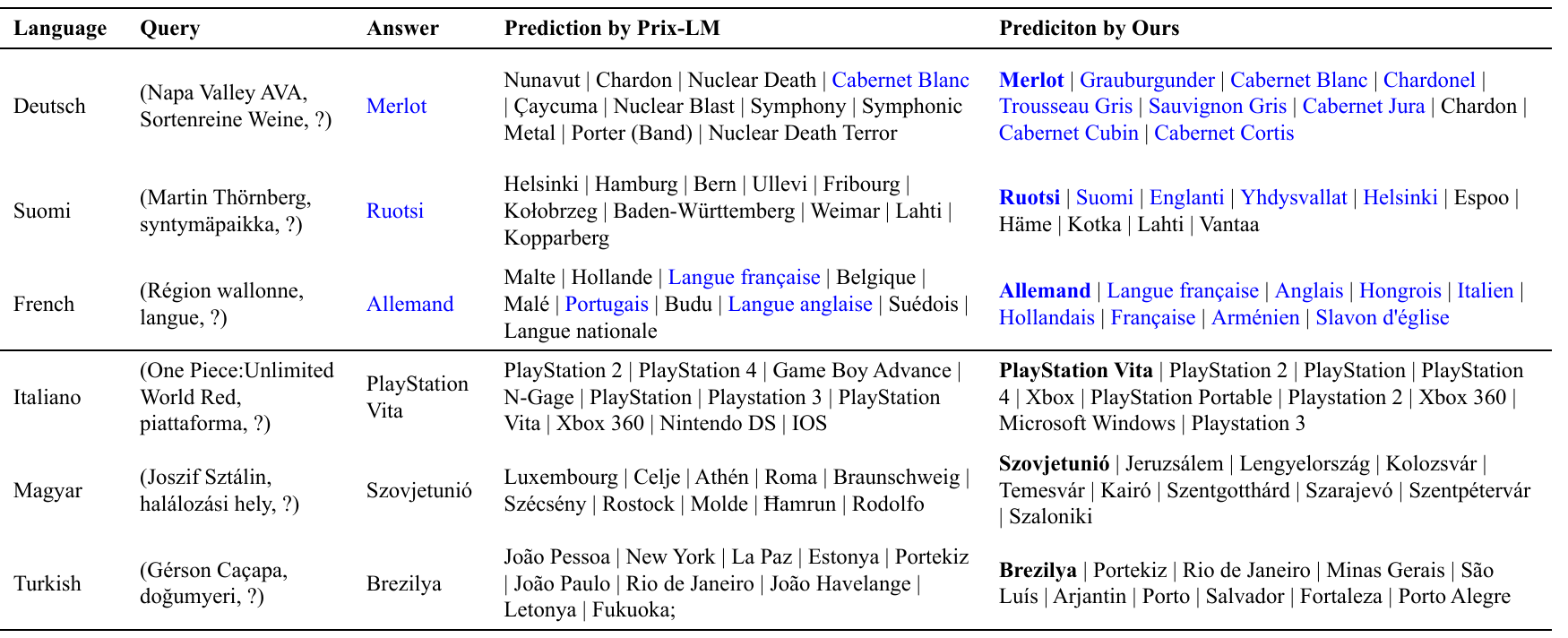}
\caption{This figure presents a comparison of the performance of our method and baseline model on a set of case studies. The \textbf{\textcolor{blue}{blue}} font is used to indicate that the predicted answer aligns with the golden answer type. The \textbf{bold} font in the predicted answer signifies the correct answer. }
\label{fig:case}
\end{figure*}

\subsection{Cross-lingual Entity Alignment}
To assess the generalizability of the proposed method, we conduct a comparison on the entity alignment task. As shown in Table~\ref{tab:ea}, we compared the proposed method with the Prix-LM. This comparison allowed us to assess the performance of the proposed method on a different task and determine its potential for use in a wider range of applications. The results show our method The results of the comparison indicate that our proposed method outperforms the Prix-LM in most of the evaluation indicators. This suggests that our method is able to generalize well to different tasks and is capable of achieving improved performance on the entity alignment task. Counterintuitively, the results show that languages with fewer resources tend to yield better performance. This may be due to the fact that the relationship between low-resource entity pairs is relatively simple and easier for the model to learn.

\subsection{Ablation Experiment}
Our proposed method introduces a novel approach for extracting both global and local knowledge through the use of a scoring function function and the maximization of mutual information. As shown in Table~\ref{tab:ablation}, we conducted an extensive comparison with various alternatives for scoring function and mutual information estimation, showcasing the superior performance of the proposed method. And we also verified the effect of different module, our findings indicate that the use of global features is better than local features, and that the difference between local and global results on the H10 metric is minimal. This supports our expectation that local features improve the ranking of entity types. Using the tasks matrix reduces some of the noise and allows faster convergence, which is the key to improving performance. We will include these results in our revised manuscript and provide a detailed discussion on their implications.

\begin{table}
\small
\centering
\setlength{\tabcolsep}{2.8mm}{
\begin{tabular}{lrrr}
\hline
Model      & Hits@1 & Hits@3 & Hits@10 \\ \hline
Prix-LM    & 19.04  & 28.58  & 37.49 \\ 
Ours & \textbf{21.39}  & \textbf{31.99}  & \textbf{43.50}   \\ \hline
w/o local  & 21.11  & 31.64  & 42.23   \\
w/o global  & 19.45  &  29.51  &  42.71   \\
w/o mask & 20.61   & 30.12  & 42.11   \\ \hline
Ours+RotatE  & 21.15  & 31.43  & 43.32   \\
Ours+ComplEx & 19.71  & 30.52  & 41.87   \\ \hline
Ours+GAN      & 20.71  & 31.36  & 43.14   \\
Ours+DV      & 18.98  & 29.61  & 41.60   \\ \hline
\end{tabular}}
\caption{This table presents the average results of seven languages of the ablation experiment which investigates the impact of different methods for acquiring global and local knowledge and different module effects. The results of the experiment provide insights into the relative importance of these different methods for improving the performance of the mKGC. }
\label{tab:ablation}
\end{table}

\subsection{Answer Length Comparison}
\begin{figure}[htp]
\centering
\includegraphics[height=6cm]{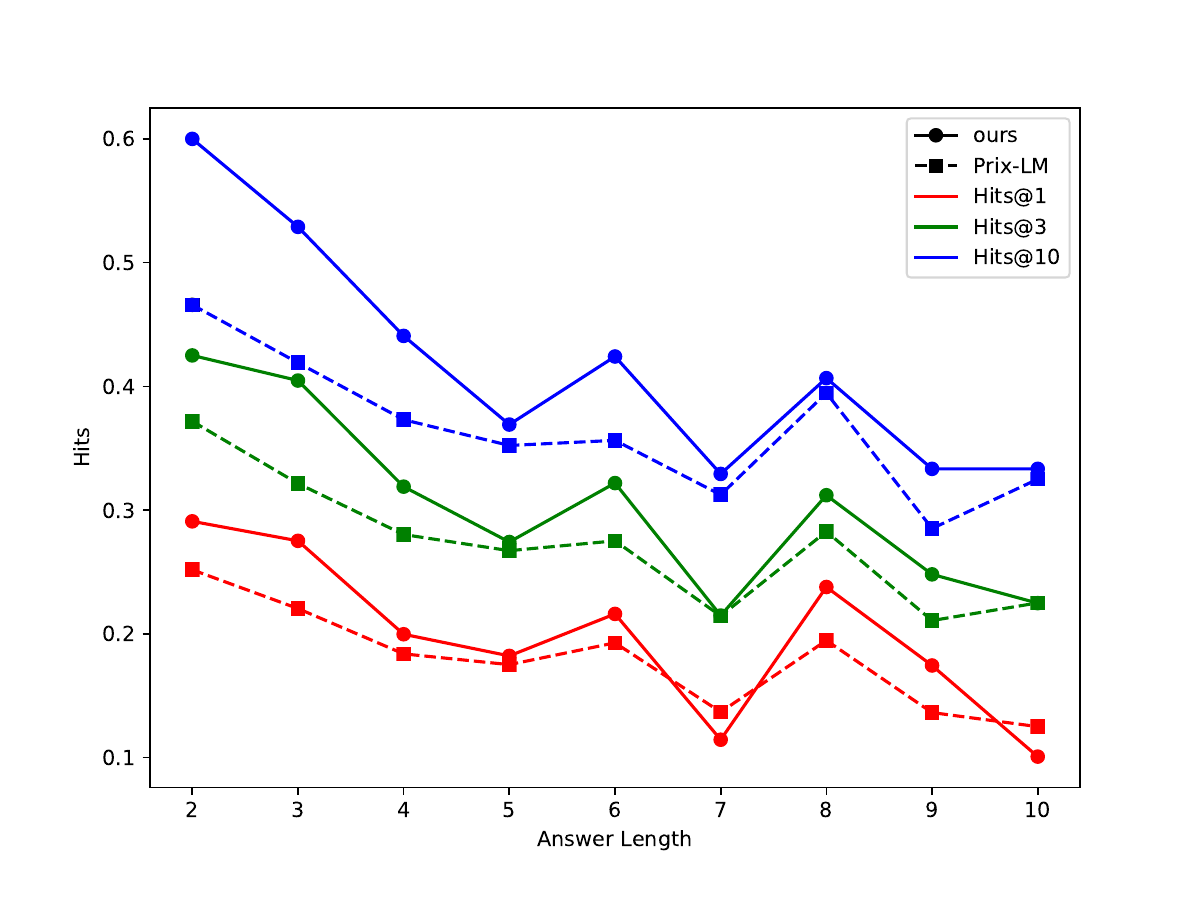}
\caption{The figure presents the results of the Hits@k evaluation metric for a mKGC task, focusing on answers of varying lengths. In order to facilitate a more straightforward analysis, the results are limited to those sets of lengths that have more than 100 occurrences.  }
\label{fig:lenth}
\end{figure}

As shown in Figure~\ref{fig:lenth}, we compared the performance of the proposed method on answers of different lengths to assess its robustness. The results of this comparison demonstrate that the proposed method exhibits strong performance across a range of answer lengths, indicating its ability to handle diverse inputs effectively. The results show that our method outperforms the baseline in terms of Hits values for answers of various lengths, with particularly strong performance on short answers.

\subsection{Case Study}
As shown in Figure~\ref{fig:case}, we compare the performance of our method with Prix-LM on a set of real examples. The predicted answers generated by both methods are presented and analyzed in order to evaluate the  effectiveness of each approach for mKGC. The results of these case studies provide additional evidence for the effectiveness of our approach in comparison to the baseline model. Our analysis of the top three cases reveals that our method produces a higher number of predictions of the same type as the correct answer compared to the baseline model. This finding suggests that our approach effectively addresses the task bias and demonstrates the adaptability of the PLM for the KGC task. Despite, the predicted answer types in the bottom three examples are all same, our method is able to accurately identify the correct answer. This demonstrates the robustness and effectiveness of our approach in generating accurate results even in situations where the predicted answers type are similar.

\section{Related Work}
\subsection{Embedding-based Methods for KGC}
There has been a amount of research focused on developing embedding-based methods for finding potential knowledge within a knowledge graph ~\cite{wang2017knowledge, dai2020survey}. These methods typically involve representing entities and relations within the graph as low-dimensional vector embeddings. Such like TransE ~\cite{bordes2013translating} makes entity and relation vectors follow the translational principle $\textbf{h} + \textbf{r} = \textbf{t}$. The choice of scoring function and the specific vector space used can have a significant impact on the performance of the method, including RotatE\cite{sun2019rotate}, TransH \cite{wang2014knowledge}, HolE\cite{nickel2016holographic}, ComplEx\cite{trouillon2016complex}. However embedding-based methods may not fully consider the background knowledge that is implicit in the text associated with entities and relations.

\subsection{Pretrained Language Models for KGC}
Recently, some research leverage pretrained language models to complete KGC task. There methods represent entities and relations by PLMs, and score high for positive triplets \cite{lv2022pre, kim2020multi}. This manner enables the introduction of knowledge that has already been learned in PLMs. 

\noindent To fully utilize the PLM, some research focus on generative paradigm for knowledge graph construction ~\cite{ye2022generative}. GenKGC ~\cite{xie2022discrimination} transforms knowledge graph completion into a sequence-to-sequence generation task base on pretrained language model and propose relation-guided demonstration and entity-aware hierarchical decoding. COMET ~\cite{bosselut-etal-2019-comet} propose the Commonsense Transformer to generate commonsense automatically. KGT5 ~\cite{saxena2022sequence} consider KG link prediction as sequence-to-sequence tasks base on a single encoder-decoder Transformer. It reduce model size for KG link prediction compare with embedding-based methods. While previous efforts to utilize PLMs for KGC have demonstrated effectiveness, they have not fully considered the inherently knowledge-based nature of KGC tasks. This oversight may hinder the full potential of such models in addressing the unique challenges and requirements of KGC.

\section{Conclusion}
Our work improve the multilingual knowledge graph completion performance base on PLM and generative paradigms. We propose two knowledgeable tasks to integrate global and local knowledge into answer generation given a query. The global knowledge improves the  type consistency  of the generated answers.  Local knowledge enhances the accuracy of answer generation. We conducted experiments and the results showed that the proposed method is better than the previous model.

\section{Limitations}
While our approach effectively predicts the relationships between entities in a knowledge graph, there are limitations in the scope of knowledge graph resources that can be modeled. The knowledge graph contains a vast array of resources, including attributes, descriptions, and images, which are not easily captured by embedding-based methods, but can be effectively modeled using PLMs. To improve the compatibility of KGC with actual needs, it is necessary to consider a broader range of data types in the knowledge graph and develop complementary methods to effectively incorporate them.

\section{Ethics Statement}
This paper proposes a method for Multilingual Knowledge Graph Completion, and the experiments are conducted on public available datasets. As a result, there is no data privacy concern. Meanwhile, this paper does not involve human annotations, and there are no related ethicalconcerns.


\bibliography{anthology,custom}
\bibliographystyle{acl_natbib}

\end{document}